\documentclass{article}
\usepackage{ijcai20}

\usepackage{times}

\usepackage{soul}
\usepackage{url}
\usepackage[hidelinks]{hyperref}
\usepackage[utf8]{inputenc}
\usepackage[small]{caption}
\usepackage{graphicx}
\usepackage{amsmath}
\usepackage{booktabs}
\usepackage{lipsum}
\usepackage{xcolor} 

\usepackage{amssymb}
\usepackage{amsfonts}
\usepackage[ruled,vlined]{algorithm2e}
\SetKwInput{KwInput}{Input}                %
\SetKwInput{KwOutput}{Output}              %

\usepackage{array}
\usepackage{multirow}

\usepackage{booktabs}
\usepackage{subcaption}
\usepackage{pifont}
\usepackage{color}
\usepackage{textcomp}

\newcommand{\ie}{\textit{i.e., }}
\newcommand{\eg}{\textit{e.g., }}

\usepackage{fancyhdr}
\title{Recent Advances in Adversarial Training for Adversarial Robustness}

\author{
Tao Bai$^1$\footnote{Contact Author}\and
Jinqi Luo$^1$\and
Jun Zhao$^{1}$\and
Bihan Wen$^{1}$\and
Qian Wang$^{2}$
\\
\affiliations
$^1$Nanyang Technological University, Singapore\\
$^2$Wuhan University, China\\
\emails
\{bait0002, luoj0021, junzhao, bihan.wen\}@ntu.edu.sg, qianwang@whu.edu.cn
}

\begin{document}
\maketitle
\begin{abstract}
Adversarial training is one of the most effective approaches to defending deep learning models against adversarial examples.
Unlike other defense strategies, adversarial training aims to enhance the robustness of models intrinsically.
During the last few years, adversarial training has been studied and discussed from various aspects.
A variety of improvements and developments of adversarial training are proposed, which were, however, neglected in existing surveys.
For the first time in this survey, we systematically review the recent progress on adversarial training for adversarial robustness with a novel taxonomy.
Then we discuss the generalization problems in adversarial training from three perspectives and highlight the challenges which are not fully tackled.
Finally, we present potential future directions.
\end{abstract}

\section{Introduction}
The adversarial vulnerability of deep neural networks has attracted significant attention in recent years. 
With slight but carefully-crafted perturbations, the perturbed natural images, namely adversarial examples~\cite{DBLP:journals/corr/SzegedyZSBEGF13}, can mislead state-of-the-art~(SOTA) classifiers to make erroneous predictions.
Besides classification, adversarial examples appear in various tasks like semantic segmentation, object detection, super-resolution, etc~(see the summary in~\cite{8611298}).
The existence of adversarial examples raises concerns from the public and motivates the proposals of defenses~\cite{DBLP:journals/corr/GoodfellowSS14,huang2015learning,DBLP:conf/iclr/MadryMSTV18}.
Naturally, the defenses also stimulate the development of stronger attacks, seeming like an arms race.

Among various existing defense strategies, Adversarial Training~(AT)~\cite{DBLP:journals/corr/GoodfellowSS14,DBLP:conf/iclr/MadryMSTV18} proves to be the most effective against adversarial attacks~\cite{pang2020bag,pmlr-v119-maini20a,schott2018towards}, receiving considerable attention from the research community.
The idea of adversarial training is straightforward: it augments training data with adversarial examples in each training loop.
Thus adversarially trained models behave more normally when facing adversarial examples than standardly trained models.
Mathematically, adversarial training is formulated as a min-max problem, searching for the best solution to the worst-case optimum.
The main challenge of adversarial training is to solve the inner maximization problem, which researchers are actively working on.
The last few years have witnessed tremendous efforts made by the research community.
The recent advances have resulted in a variety of new techniques in the literature, which no doubt deserves a comprehensive review.

To our best knowledge, surveys focusing on adversarial training do not exist so far, and some recent surveys~\cite{8611298,silva2020opportunities} mainly summarize existing adversarial attacks and defense methods.
Our goal in this paper is to give a brief overview of adversarial training.
We believe that this survey can provide up-to-date findings and developments happening on adversarial training. 
Notably, we carefully review and analyze adversarial training with a novel taxonomy, uniquely discuss the poor generalization ability from different perspectives, and present future research directions.
\section{Preliminaries}
\noindent \textbf{Adversarial Attacks.}
Adversarial attacks refer to finding adversarial examples for well-trained models.
In this paper, we consider only the situation where the training/test data are initially from the same distribution.
Taking classification as an example, we use $f(x; \theta): \mathbb{R}^{h\times w \times c} \rightarrow\{1 \dots k\}$ to denote an image classifier that maps an input image $x$ to a discrete label set $C$ with $k$ classes, in which $\theta$ indicates the parameters of $f$, and $h, w, c$ represent image height, width and channel, respectively.
Given the perturbation budget $\epsilon$, the adversary tries to find a perturbation $\delta \in \mathbb{R}^{h\times w \times c}$ to maximize the loss function, \eg cross-entropy loss $\mathcal{L}_{ce}$, so that $f(x+\delta)\neq f(x)$. 
Therefore, $\delta$ is estimated as
\begin{equation}\begin{array}{c}
\delta^*:=\mathop{\arg\max}\limits_{|\delta|_p \leq \epsilon} \mathcal{L}_{ce}(\theta, x+\delta, y),
\end{array}\end{equation}
where $y$ is the label of $x$, and $p$ can be 0, 1, 2 and $\infty$.
In most cases, $\epsilon$ is small so that the perturbations are imperceptible to human eyes.
Note that we only consider the $l_p$-based attacks for classification in this paper.

The adversarial counterpart $x'$ of $x$ is expressed as 
\begin{equation}\begin{array}{c}
x' := x + \delta^*.
\end{array}\end{equation}
There are some common attacks following this formulation, such as Fast Gradient Sign Method (FGSM)~\cite{DBLP:journals/corr/GoodfellowSS14}, iterative FGSM~\cite{DBLP:journals/corr/KurakinGB16}, and Projected Gradient Descent~(PGD) attack~\cite{DBLP:conf/iclr/MadryMSTV18}.

\noindent \textbf{Adversarial Robustness.}
In general, adversarial robustness is the model's performance on test data with adversarial attacks, \ie adversarial test accuracy of classification.
We expect to know the model's performance in the worst-case scenario, so the adversary should be strong enough and launch attacks in white-box settings, \ie the adversary has full knowledge of the model, such as architectures, parameters, training data, etc.
In practice, PGD attack~\cite{DBLP:conf/iclr/MadryMSTV18} is commonly employed for evaluation because of its attack effectiveness in white-box settings.

\section{Adversarial Training for Adversarial Robustness}
Currently, adversarial training is widely accepted as the most effective method in practice to improve the adversarial robustness of deep learning models~\cite{athalye2018robustness}.
However, it is still a long way to go for adversarial training to handle adversarial attacks perfectly.
Prevailing adversarial training methods~\cite{DBLP:conf/iclr/MadryMSTV18} can produce a robust model with worst-case accuracy of around 90\% on MNIST.
For slightly more challenging datasets, \eg CIFAR-10, adversarial training only achieves around 45\% and 40\% on SVHN~\cite{buckman2018thermometer}, which is far from satisfactory.
Additionally, adversarial training leads to the degraded generalization ability of deep learning models.
In this section, we first review the development of adversarial training, summarize the recent advances with a novel taxonomy, and discuss the generalization problem of adversarial training.

\begin{table}[]
\resizebox{\columnwidth}{!}{%
\begin{tabular}{ccccccc}
\toprule[1pt]
Taxonomy & Publication                              & Model Architecture          & Attack   & $\epsilon$ & Dataset       & Accuracy \\ \midrule
\multirow{5}{*}{\rotatebox{45}{\shortstack{Adversarial \\ Regularization}}} & \cite{NIPS2019_9534}                    & ResNet-152         & PGD$_{50}$   & 4/255   & ImageNet      & 47.00\% \\ \cline{2-7} 
                                    & \cite{pmlr-v97-zhang19p}                 & Wide ResNet         & CW$_{10}$    & 0.031/1 & CIFAR-10      & 84.03\% \\ \cline{2-7} 
                                    & \cite{wang2020improving}                 & ResNet-18           & PGD$_{20}$   & 8/255   & CIFAR-10      & 55.45\% \\ \cline{2-7} 
                                    & \cite{DBLP:journals/corr/abs-1803-06373} & InceptionV3         & PGD$_{10}$   & 16/255  & ImageNet      & 27.90\% \\ \cline{2-7} 
                                    & \cite{NIPS2019_8339}                    & Wide ResNet         & PGD$_{20}$   & 8/255   & CIFAR-10      & 50.03\% \\ \midrule[0.7pt]
\multirow{3}{*}{\rotatebox{45}{\shortstack{Curriculum}}}         & \cite{pmlr-v119-zhang20z}                & Wide ResNet         & PGD$_{20}$   & 16/255  & CIFAR-10      & 49.86\% \\ \cline{2-7} 
                                    & \cite{ijcai2018-520}                     & DenseNet-161        & PGD$_{7}$    & 8/255   & CIFAR-10      & 69.27\% \\ \cline{2-7} 
                                    & \cite{DBLP:conf/icml/WangM0YZG19}        & 8-Layer ConvNet     & PGD$_{20}$   & 8/255   & CIFAR-10      & 42.40\% \\ \midrule[0.7pt]
\multirow{3}{*}{\rotatebox{45}{\shortstack{Ensemble}}}           & \cite{DBLP:conf/icml/PangXDCZ19}         & Wide ResNet         & PGD$_{10}$   & 0.005   & CIFAR-100     & 32.10\% \\ \cline{2-7} 
                                    & \cite{kariyappa2019improving}            & ResNet-20           & PGD$_{30}$   & 0.09/1  & CIFAR-10      & 46.30\% \\ \cline{2-7} 
                                    & \cite{NEURIPS2020_3ad7c2eb}             & ResNet-20           & PGD$_{20}$   & 0.01/1  & CIFAR-10      & 52.4\%  \\ \midrule[0.7pt]
\multirow{3}{*}{\rotatebox{45}{\shortstack{Adaptive \\ $\epsilon$}}}           & \cite{balaji2019instance}                & ResNet-152          & PGD$_{1000}$ & 8/255   & ImageNet      & 59.28\% \\ \cline{2-7} 
                                    & \cite{Ding2020MMA}                       & Wide ResNet         & PGD$_{100}$  & 8/255   & CIFAR-10      & 47.18\% \\ \cline{2-7} 
                                    & \cite{cheng2020cat}                      & Wide ResNet         & PGD$_{20}$   & 8/255   & CIFAR-10      & 73.38\% \\ \midrule[0.7pt]
\multirow{4}{*}{\rotatebox{45}{\shortstack{Semi-\\Unsupervised}}}  & \cite{NIPS2019_9388}                    & Wide ResNet         & FGSM         & 8/255   & CIFAR-10      & 62.18\% \\ \cline{2-7} 
                                    & \cite{NIPS2019_9298}                    & Wide ResNet         & PGD$_{10}$   & 8/255   & CIFAR-10      & 63.10\% \\ \cline{2-7} 
                                    & \cite{DBLP:journals/corr/abs-1906-00555} & Customized ResNet   & PGD$_{7}$    & 8/255   & CIFAR-10      & 42.48\% \\ \cline{2-7} 
                                    & \cite{DBLP:journals/corr/abs-1906-12340} & Wide ResNet         & PGD$_{20}$   & 0.3/1   & ImageNet      & 50.40\% \\ \midrule[0.7pt]
\multirow{9}{*}{\rotatebox{45}{Efficient}}          & \cite{NIPS2019_8597}                    & Wide ResNet         & PGD$_{100}$  & 8/255   & CIFAR-10      & 46.19\% \\ \cline{2-7} 
                                    & \cite{Wong2020FastIB}                    & ResNet-50           & PGD$_{40}$   & 2/255   & ImageNet      & 43.43\% \\ \cline{2-7} 
                                    & \cite{Andriushchenko2020UnderstandingAI} & ResNet-50           & PGD$_{50}$   & 2/255   & ImageNet      & 41.40\% \\ \cline{2-7} 
                                    & \cite{kim2020understanding}              & PreActResNet-18     & FGSM       & 8/255   & CIFAR-10      & 50.50\% \\ \cline{2-7} 
                                    & \cite{9157154}                           & Wide ResNet         & PGD$_{40}$   & 8/255   & MNIST         & 88.51\% \\ \cline{2-7} 
                                    & \cite{song2018improving}                 & Customized ConvNet  & PGD$_{20}$   & 4/255   & CIFAR-10      & 58.10\% \\ \cline{2-7} 
                                    & \cite{vivek2020regularizers}             & Wide ResNet         & PGD$_{100}$  & 0.3/1   & MNIST         & 90.03\% \\ \cline{2-7} 
                                    & \cite{huang2020bridging}                 & Wide ResNet         & PGD$_{20}$   & 8/255   & CIFAR-10      & 45.80\% \\ \cline{2-7} 
                                    & \cite{Zhang2019YouOP}                    & Wide ResNet         & PGD$_{20}$   & 8/255   & CIFAR-10      & 47.98\% \\ \midrule[0.7pt]
\multirow{6}{*}{\rotatebox{45}{Others}}              & \cite{DBLP:conf/nips/DongDP0020}         & Wide ResNet         & PGD$_{20}$   & 8/255   & CIFAR-100     & 29.40\% \\ \cline{2-7} 
                                    & \cite{Wang_2019_ICCV}                  & Wide ResNet         & CW$_{200}$   & 4/255   & CIFAR-10      & 60.30\% \\ \cline{2-7} 
                                    & \cite{NIPS2019_8459}                    & Wide ResNet         & PGD$_{20}$   & 8/255   & CIFAR-100     & 47.20\% \\ \cline{2-7} 
                                    & \cite{pang2020boosting}                  & Wide ResNet         & PGD$_{500}$  & 8/255   & CIFAR-10      & 60.75\% \\ \cline{2-7} 
                                    & \cite{Lee_2020_CVPR}                   & PreActResNet-18     & PGD$_{20}$   & 8/255   & Tiny ImageNet & 20.31\% \\ \cline{2-7} 
                                    & \cite{zhang2020adversarial}              & Wide ResNet         & PGD$_{20}$   & 8/255   & CIFAR-10      & 45.11\% \\ \midrule[0.7pt]
\multirow{1}{*}{Benchmark}     & \cite{DBLP:conf/iclr/MadryMSTV18}        & ResNet-50           & PGD$_{20}$   & 8/255   & CIFAR-10      & 45.80\% \\  \bottomrule[1pt]
\end{tabular}%
}
\caption{A summary of experimental results for various adversarial training methods. All the attacks are under $l_{\infty}$ norm.}
\label{tab:adv training}
\end{table}

\subsection{The Origin of Adversarial Training}
The initial idea of adversarial training is first brought to light by~\cite{DBLP:journals/corr/SzegedyZSBEGF13}, where neural networks are trained on a mixture of adversarial examples and clean examples.
\citeauthor{DBLP:journals/corr/GoodfellowSS14}~(\citeyear{DBLP:journals/corr/GoodfellowSS14}) went further and proposed FGSM to produce adversarial examples during training.
Yet, their trained models remain vulnerable to iterative attacks~\cite{DBLP:conf/iclr/TramerKPGBM18} as these approaches utilized a linear function to approximate the loss function, leading to sharp curvature near data points on the decision surface of the corresponding deep models.
The existence of sharp curvature is also known as gradient masking~\cite{Papernot2017Practical}.

Unlike the prior works in that models are trained on a mixture of clean data and adversarial data, a line of research trains models with adversarial data only.
For the first time, \citeauthor{huang2015learning}~(\citeyear{huang2015learning}) defined a min-max problem that the training procedure is forced to minimize classification error against an adversary who perturbs the input and maximizes the classification error. 
They also pointed out that the key to solving this min-max problem is finding strong adversarial examples.
\citeauthor{shaham2018understanding}~(\citeyear{shaham2018understanding}) considered this min-max problem from a robust optimization perspective and proposed the framework of adversarial training.
The formulation is illustrated below:
\begin{equation}\begin{array}{c}
\min \limits_{\theta} \mathbb{E}_{(x, y) \sim \mathcal{D}}\left[\max \limits_{\delta \in B(x,\varepsilon)} \mathcal{L}_{ce}(\theta, x+\delta, y)\right],
\end{array}
\label{eqn:adv_training}
\end{equation}
where $(x, y) \sim \mathcal{D}$ represents training data sampled from distribution $\mathcal{D}$ and $B(x,\varepsilon)$ is the allowed perturbation set, expressed as
$B(x,\varepsilon):=\left\{x+\delta \in \mathbb{R}^{m}  \mid  \|\delta\| _{p} \leq \varepsilon\right\}$.
\citeauthor{DBLP:conf/iclr/MadryMSTV18}~(\citeyear{DBLP:conf/iclr/MadryMSTV18}) gave a reasonable interpretation of this formulation: the inner maximization problem is finding the worst-case samples for the given model, and the outer minimization problem is to train a model robust to adversarial examples.

With such connection, \citeauthor{DBLP:conf/iclr/MadryMSTV18}~(\citeyear{DBLP:conf/iclr/MadryMSTV18}) employed a multi-step gradient based attack known as PGD attack for solving the inner problem as follows:
\begin{equation}
x^{t+1}=\operatorname{Proj}_{x+B(x,\varepsilon)}\left(x^{t}+\alpha \operatorname{sign}\left(\nabla_{x^{t}} \mathcal{L}_{ce}\left(\theta, x^{t}, y\right)\right)\right),
\label{eqn:pgd}
\end{equation}
where $t$ is the current step and $\alpha$ is the step size.
Further, they investigated the inner maximization problem from the landscape of adversarial examples and gave both theoretical and empirical proofs of local maxima's tractability with PGD.
Through extensive experiments, their approach (PGD-AT) significantly increased the adversarial robustness of deep learning models against a wide range of attacks, which is a milestone of adversarial training methods.
As most derivative works followed their designs and settings, PGD-AT became a critical benchmark and is regarded as the standard way to do adversarial training in practice.

\subsection{Taxonomy of Adversarial Training}\label{subsec:advances}
In this subsection, we review the recent advances of adversarial training in last few years, categorized by different understandings on adversarial training.
A summary of selected adversarial training methods is provided in Table~\ref{tab:adv training}.

\subsubsection{Adversarial Regularization}
The idea of adversarial regularization first appears in~\cite{DBLP:journals/corr/GoodfellowSS14}.
Besides cross-entropy loss, they added a regularization term in the objective function, which is based on FGSM and expressed as $\mathcal{L}\left(\boldsymbol{\theta}, x+\epsilon \operatorname{sign}\left(\nabla_{x} \mathcal{L}(\boldsymbol{\theta}, x, y)\right)\right.$.
\citeauthor{DBLP:conf/iclr/KurakinGB17}~(\citeyear{DBLP:conf/iclr/KurakinGB17}) extended this FGSM-based regularization term by controlling the ratio of adversarial examples in batches so that it can scale up to ImageNet.
Their methods' effectiveness is validated on single-step attacks as they believe the linearity of neural networks is attributed to the existence of adversarial examples~\cite{DBLP:journals/corr/GoodfellowSS14}.
However, \citeauthor{NIPS2019_9534}~(\citeyear{NIPS2019_9534}) calculated the absolute difference between the adversarial loss and its first-order Taylor expansion, concluding that more robust models usually have smaller values of local linearity.
Correspondingly, they replaced the FGSM-based regularization with a Local Linearity Regularization for adversarial robustness.

Distinct from previous methods, \cite{pmlr-v97-zhang19p} decomposed the robust error $\mathcal{R}_{\mathrm{rob}}$ as the sum of natural error $\mathcal{R}_{\mathrm{nat}}$ and boundary error $\mathcal{R}_{\mathrm{db}}$.
Boundary error occurs when the distance between data and the decision boundary is sufficiently small~(less than $\epsilon$), which is also the reason for adversarial examples' existence.
So they proposed TRADES to minimize the $\mathcal{R}_{\mathrm{db}}$ by solving the following problem:
\begin{equation}
\min _{f} \mathbb{E}\{{\mathcal{L}(f(x), y)} +{\max _{x^{\prime} \in \mathbb{B}(x, \epsilon)} \mathcal{L}\left(f(x),f\left(x^{\prime}\right)\right)/ \lambda}\},
\end{equation}
where $\lambda$ is a coefficient determining the strength of regularization.
Such decomposition is proved to be effective, and TRADES outperforms PGD-AT on CIFAR-10 with error rates reduced by 10\%.
One problem of TRADES is that the regularization term is designed to push natural examples and their adversarial counterparts together, no matter natural data are classified correctly or not.
\cite{wang2020improving} investigated the influence of misclassified examples and proposed Misclassification Aware adveRsarial Training (MART), which emphasizes on misclassified examples with weights $1-\mathcal{P}_{y}\left(x, \theta\right)$, where $\mathcal{P}_{y}\left(x, \theta\right)$ is the probability of ground truth label $y$.

Due to the amplification of deep models, imperceptible noises could lead to substantial changes in feature space~\cite{DBLP:journals/corr/GoodfellowSS14}.
Some works analyze adversarial training from the perspective of representation.
\citeauthor{DBLP:journals/corr/abs-1803-06373}~(\citeyear{DBLP:journals/corr/abs-1803-06373}) proposed Adversarial Logit Pairing (ALP), encouraging logits for pairs of examples to be similar.
But ALP initially is not useful due to the wrong formulation of adversarial training objectives~\cite{engstrom2018evaluating}.
Further to enhance the alignment of representations of natural data and their adversarial counterparts, \citeauthor{NIPS2019_8339}~(\citeyear{NIPS2019_8339}) adopted the prevalent triplet loss for regularization, which uses adversarial examples as anchors.

Adversarial regularization is an essential variant of adversarial training~\cite{shaham2018understanding}.
Compared to the original formulation of adversarial training, adversarial regularization is more flexible and requires a deep understanding of adversarial robustness.
Also, the decomposition of robust error indeed paves the way for unlabeled data to enhance adversarial robustness.

\subsubsection{Curriculum-based Adversarial Training}
According to the formulation of adversarial training, the inner problem is always trying to find the worst-case samples.
One natural question is: are those worst-case samples always suitable for adversarial training?
\citeauthor{pmlr-v119-zhang20z}~(\citeyear{pmlr-v119-zhang20z}) found that adversarial examples generated by strong attacks significantly cross over the decision boundary and are close to natural data.
As PGD-AT only utilizes adversarial examples for training, this leads to overfitting of adversarial examples~\cite{ijcai2018-520}.

For alleviating the overfitting, researchers adapt the idea of curriculum training to adversarial training.
\citeauthor{ijcai2018-520}~(\citeyear{ijcai2018-520}) proposed Curriculum Adversarial Training (CAT), with an assumption that PGD with more steps generates stronger adversarial examples.
Starting from a small number of steps, CAT gradually increases the iteration steps of PGD until the model achieves a high accuracy against the current attack.
Different from CAT, Friendly Adversarial Training~(FAT)~\cite{pmlr-v119-zhang20z} adapts early stopping when performing PGD attacks and returns adversarial data near the decision boundary for training.
Both CAT and FAT adjust the attacks' strength in a practical way, where a quantitative criterion is missing.
From the convergence point of view, \citeauthor{DBLP:conf/icml/WangM0YZG19}~(\citeyear{DBLP:conf/icml/WangM0YZG19}) designed First-Order Stationary Condition~(FOSC) to estimate the convergence quality of the inner maximization problem. 
The closer the FOSC to 0, the stronger the attack.

Such curriculum-based methods help improve the generalization of clean data while preserving adversarial robustness.
One possible reason for their success is weak attacks in early training stages are associated with generalization~\cite{DBLP:conf/icml/WangM0YZG19}.
In addition to relieving overfitting, curriculum-based methods reduce training time due to the varying iteration numbers of PGD for solving the inner maximization problem.

\subsubsection{Ensemble Adversarial Training}\label{subsubsection:ensemble}
\citeauthor{DBLP:conf/iclr/TramerKPGBM18}~(\citeyear{DBLP:conf/iclr/TramerKPGBM18}) firstly introduced ensemble learning into adversarial training, called Ensemble Adversarial Training~(EAT), where training data is augmented with adversarial examples generated from different target models instead of a single model.
The advantage of EAT is that it helps alleviate the sharp curvatures caused by the single-step attacks \eg FGSM.
However, the interaction among different target models is neglected~\cite{DBLP:conf/iclr/TramerKPGBM18}. 
Specifically, standardly trained target models may have similar predictions or representations~\cite{dauphin2014identifying} and share the adversarial subspace~\cite{tramer2017space}, which potentially hurts the performance of EAT.

For promoting the diversity among target models, several improvements are proposed, such as the adaptive diversity promoting regularizer~\cite{DBLP:conf/icml/PangXDCZ19}, forcing different models to be diverse in non-maximal predictions;
maximizing the cosine distances among each target models' input gradients~\cite{kariyappa2019improving} (input gradients refer to the gradients of the loss function w.r.t. the input); 
and maximizing the vulnerability diversity~\cite{NEURIPS2020_3ad7c2eb}, which is defined as the sum of losses for two models with crafted images containing non-robust features~\cite{ilyas2019adversarial}.

Intrinsically, such ensemble methods are useful for approximating the optimal value of the inner maximization problem in adversarial training.
As proved in~\cite{DBLP:conf/iclr/TramerKPGBM18}, models trained with EAT have better generalization abilities regardless of the perturbation types.
To conclude, adding the number and diversity of target models in training is a practical and useful way to approximate the space of adversarial examples, which is challenging to be described explicitly.

\subsubsection{Adversarial Training with Adaptive \texorpdfstring{$\epsilon$}{}}\label{subsubsec:adv_adapt}
As shown in Equation~(\ref{eqn:adv_training}), the parameters of attacks are predefined and fixed during training, such as $\epsilon$.
Some works~\cite{balaji2019instance,Ding2020MMA} argued individual data points might have different intrinsic robustness, \ie different distances to the classifier's decision boundary; however, adversarial training with fixed $\epsilon$ treats all data equally.

Considering the individual characteristic of adversarial robustness, researchers propose to do adversarial training at the instance level.
\citeauthor{balaji2019instance}~(\citeyear{balaji2019instance}) firstly presented Instance Adaptive Adversarial Training~(IAAT), where $\epsilon$ is selected to be as large as possible, ensuring images within $\epsilon$-ball of $x$ are from the same class.
This strategy helps IAAT relieve the trade-off between robustness and accuracy, though there is a slight drop in robustness.
Unlike IAAT, another work called Margin Maximization Adversarial Training (MMA)~\cite{Ding2020MMA} directly maximizes the margin-distances between data points and the model's decision boundary, which is estimated by the adversarial perturbations with the least magnitudes.
The manner of choosing $\epsilon$ in MMA is more reasonable as $\epsilon$ is sufficiently small, and such small $\epsilon$ in spatial domain hardly changes the classes of images substantially, especially for high-resolution images.
The following work, Customized Adversarial Training~(CAT)~\cite{cheng2020cat} further applies adaptive label uncertainty to prevent over-confident predictions based on adaptive $\epsilon$. 

Adversarial training with adaptive $\epsilon$ is a good exploration.
However, empirical evidence shows many standard datasets are distributionally separated, \ie the distances inter classes are larger than $\epsilon$ used for attacks~\cite{yang2020closer}.
This reflects the limitation of current adversarial training methods on finding proper decision boundaries.

\subsubsection{Adversarial Training with Semi/Unsupervised Learning}\label{subsubsec:adv_unsupervised}
One key observation in supervised adversarial training methods~\cite{DBLP:conf/iclr/MadryMSTV18,pmlr-v97-zhang19p} is adversarial accuracy in testing is much lower than in training.
There is a large generalization gap in adversarial training (see Figure 1 in~\cite{NIPS2018_7749}).
The recent work~\cite{NIPS2018_7749} studied this problem from the perspective of sample complexity.
It is theoretically proved that adversarially robust training requires substantially larger datasets than standard training.
However, quality datasets with labels are expensive to collect, which is of particular interest in practice.
Alternatively, several works appeared concurrently, exploring the possibility of training with additional unlabeled data.

Following the analysis of Gaussian models in~\cite{NIPS2018_7749}, a couple of works~\cite{NIPS2019_9388,NIPS2019_9298,DBLP:journals/corr/abs-1906-00555} theoretically show that unlabeled data significantly reduces the sample complexity gap between standard training and adversarial training.
They share the same idea of decomposing the adversarial robustness like TRADES and utilize unlabeled data for stability while labeled data for classification.
Empirically, they investigated the impact of different factors on adversarial training like label noise, distribution shift, and the amount of additional data.
On the other hand, \citeauthor{NIPS2019_8792}~(\citeyear{NIPS2019_8792}) introduced some new complexity measures like Adversarial Rademacher Complexity and Minimum Supervision Ratio for theoretical analysis on generalization.
It is also observed that adversarial robustness is benefited by self-supervised training~\cite{DBLP:journals/corr/abs-1906-12340}.

It is inspiring to see the improvement of adversarial robustness brought by additional unlabeled data. 
However, theoretical or empirical guarantees that how much additional data are needed precisely still lack.
Besides, the cost of such methods should not be neglected, including collecting data and training adversarially on data multiple times larger than original datasets.

\subsubsection{Efficient Adversarial Training}
One well-known limitation of conventional adversarial training methods like PGD-AT is that they take 3-30 times longer than standard training before the model converges~\cite{NIPS2019_8597}. 
The main reason is the min-max problem described in Equation~(\ref{eqn:adv_training}) is solved iteratively.
This line of research of adversarial training aims to reduce the time cost while keeping the performances of adversarial training.

As the first attempt, the core idea of free adversarial training~(Free-AT)~\cite{NIPS2019_8597} is to reuse the gradients computed in the backward pass when doing forward pass.
In Free-AT, both the model parameters and image perturbations are updated simultaneously.
Concretely, for the same mini-batch data, the same operation is done for $m$ times in a row, equivalent to utilizing strong adversarial examples in PGD-AT.
Further upon Free-AT, \citeauthor{Wong2020FastIB}~(\citeyear{Wong2020FastIB}) proposed fast adversarial training~(FAST-AT), which utilizes FGSM with random initialization and is as effective as the PGD-AT.
They also attributed the failure of FGSM-based adversarial training methods to the \textit{catastrophic overfitting} and zero-initialized perturbation.
However, \citeauthor{Andriushchenko2020UnderstandingAI}~(\citeyear{Andriushchenko2020UnderstandingAI}) found these fast training methods~\cite{NIPS2019_8597,Wong2020FastIB} suffer from catastrophic overfitting as well.
They also pointed out the reason for randomization to take effect in~\cite{Wong2020FastIB} is that randomization slightly reduces the magnitude of perturbations.
\citeauthor{kim2020understanding}~(\citeyear{kim2020understanding}) supported the above finding and demonstrated catastrophic overfitting is because single-step adversarial training uses only adversarial examples with maximum perturbations.
For the purpose of preventing catastrophic overfitting, many improvements are proposed like  GradAlign~\cite{Andriushchenko2020UnderstandingAI}, dynamic schedule~\cite{9157154}, inner interval verification~\cite{kim2020understanding}, domain adaption~\cite{song2018improving} and regularization methods~\cite{vivek2020regularizers,huang2020bridging}.

Intrinsic from the above works, \citeauthor{Zhang2019YouOP}~(\citeyear{Zhang2019YouOP}) proposed You Only Propagate Once~(YOPO) from the perspective of Pontryagin’s Maximum Principle.
According to their analysis on adversarial training, they observed that adversarial gradients update is only related to the first layer of neural networks.
This property enables YOPO to focus on the first layer of the proposed network architecture for adversary computation while other layers are frozen, significantly reducing the numbers of forward and backward propagation.
The authors claimed that Free-AT is a particular case of YOPO.

\subsubsection{Other Variants}
In addition to the above branches of adversarial training methods, several other variants of adversarial training are summarized as follows.
Some works modify the learning objectives of vanilla adversarial training, like adversarial distributional training~\cite{DBLP:conf/nips/DongDP0020} where a distribution-based min-max problem is derived from a general view;
bilateral adversarial training~\cite{Wang_2019_ICCV} where the model is training on both perturbed images and labels; 
and adversarial training based on feature scatter~\cite{NIPS2019_8459}, which utilizes a distance metric for sets of natural data and their counterparts and produces adversarial examples in feature space.
Some replace the fundamental components of models for better performances, like hypersphere embedding~\cite{pang2020boosting}, and smoothed ReLU function~\cite{xie2020smooth}.
Last, some propose to augment adversarial examples by interpolation, such as AVmixup~\cite{Lee_2020_CVPR} and adversarial interpolation training~\cite{zhang2020adversarial}.

\subsection{Generalization Problem in Adversarial Training}
For deep learning algorithms, generalization is a significant characteristic.
Though most efforts in this research community are paid for improving adversarial training under given adversarial attacks, the voice of discussions on generalization is getting louder.
This subsection mainly reviews the studies on the generalization of adversarial training from three aspects: standard generalization, adversarially robust generalization, and generalization on unseen attacks.

\subsubsection{Standard Generalization}
Despite the success in improving the robustness of neural networks to adversarial attacks, adversarial training is observed that hurts standard accuracy badly~\cite{DBLP:conf/iclr/MadryMSTV18}, leading to the discussion of relationships between adversarial robustness and standard accuracy. 
We refer to it as standard generalization.

One popular viewpoint is the trade-off between adversarial robustness and standard accuracy.
\citeauthor{tsipras2018robustness}~(\citeyear{tsipras2018robustness}) claimed that standard accuracy and adversarial robustness might at odds and demonstrated the existence of the trade-off via a binary classification task.
\citeauthor{DBLP:journals/corr/abs-1808-01688}~(\citeyear{DBLP:journals/corr/abs-1808-01688}) evaluated the recent SOTA ImageNet-based models on multiple robustness metrics. 
They concluded a linearly negative correlation between the logarithm of model classification accuracy and model robustness. 
\citeauthor{pmlr-v97-zhang19p}~(\citeyear{pmlr-v97-zhang19p}) decomposed the robust error as the sum of natural error and the boundary error and provided a tight upper bound for them, theoretically characterizing the trade-off.

However, some works have different opinions that adversarial robustness and standard accuracy are not opposing.
\citeauthor{Stutz_2019_CVPR}~(\citeyear{Stutz_2019_CVPR}) studied the manifold of adversarial examples and natural data.
They confirmed the existence of adversarial examples on the manifold of natural data, adversarial robustness on which is equivalent to generalization.
\citeauthor{yang2020closer}~(\citeyear{yang2020closer}) investigated various datasets, from MNIST to Restricted ImageNet, showing these datasets are distributionally separated, and the separation is usually larger than $2\epsilon$ (the value of $\epsilon$ differs in different datasets, \eg 0.1/1 for MNIST, 8/255 for CIFAR).
It indicates the existence of robust and accurate classifiers.
They also claimed existing training methods fail to impose local Lipschitzness or are insufficiently generalized.
Experiments in~\cite{raghunathan2019adversarial} support this statement, where additional unlabeled data is proved to help mitigate the trade-off.

As suggested in~\cite{yang2020closer}, the trade-off might not be inherent but a consequence of current adversarial training methods.
Though researchers haven't reached a consensus on the cause of the trade-off, existing evidence does reveal some limitations on adversarial training.
Adversarial robustness should not be at the cost of standard accuracy.
Some variants of adversarial training show better standard generalization empirically, such as adaptive $\epsilon$ for adversarial training reviewed in Section~\ref{subsubsec:adv_adapt}, robust local features~\cite{song2020robust} and $\mathcal{L}_1$ penalty~\cite{xing2020generalization}.

\subsubsection{Adversarially Robust Generalization}
The phenomenon that adversarially trained models do not perform well on adversarially perturbed test data is firstly observed in~\cite{DBLP:conf/iclr/MadryMSTV18}.
In other words, there is a large gap between the training accuracy and test accuracy on adversarial data.
Other than CIFAR-10, similar experimental results are observed on multiple datasets, such as SVHN, CIFAR-100, and ImageNet~\cite{rice2020overfitting}.
These gaps indicate that a severe overfitting happens in current adversarial training methods.
Such overfitting is initially studied by \cite{NIPS2018_7749}, who refer to it as adversarially robust generalization.

\citeauthor{NIPS2018_7749}~(\citeyear{NIPS2018_7749}) revealed the difficulty of obtaining a robust model with the fact that more training data are required for adversarially robust generalization.
Later many efforts are made to improve generalization empirically, such as adversarial training with semi/unsupervised Learning, AVmixup, and robust local feature~\cite{song2020robust}.
In contrast, \citeauthor{rice2020overfitting}~(\citeyear{rice2020overfitting}) systematically investigated various techniques used in deep learning like $\ell_{1}$ and $\ell_{2}$ regularization, cutout, mixup and early stopping, where \textbf{early stopping} is found to be the most effective and confirmed by~\cite{pang2020bag}.

On the other hand, though researchers attempt to analyze this generalization problem with different tools like Rademacher complexity~\cite{DBLP:conf/icml/YinRB19} and VC dimention~\cite{cullina2018pac}, theoretical progress is, in fact, limited, and the generalization problem is far from being solved.

\subsubsection{Generalization on Unseen Attacks}
The last significant property of adversarial training is to generalize on unseen attacks.
It is proved that the specific type of attacks are not sufficient to represent the space of possible perturbations~\cite{DBLP:conf/iclr/TramerKPGBM18,DBLP:journals/corr/GoodfellowSS14}.
However, in adversarial training, the inner maximization problem's constraints: $l_p$ norm and $\epsilon$ are pre-fixed.
Thus, adversarially trained models, which are robust to a specific attack, \eg $l_{\infty}$ adversarial examples, can be circumvented easily by different types of attacks, \eg other $l_p$ norms, or larger $\epsilon$, or different target models~\cite{kang2019transfer}.
Simply combining perturbations with different $l_p$ norms in adversarial training proves to be useless as well~\cite{NEURIPS2019_5d4ae76f}.
This kind of poor generalization to other attacks significantly degrades the reliability of adversarial training.

Such limitation is intrinsically caused by adversarial training itself, and the key is how to solve the inner problem properly.
The research line in EAT can be seen as the first attempt to approximate the optimal solutions to the inner problem by increasing the number and diversity of targeted models during training.
Similarly, \citeauthor{pmlr-v119-maini20a}~(\citeyear{pmlr-v119-maini20a}) adopted adversaries under different $l_p$ norm and use the steepest descent to approximate the optimal values for the inner problem.
\citeauthor{DBLP:conf/nips/DongDP0020}~(\citeyear{DBLP:conf/nips/DongDP0020}) proposed to explicitly model the distribution of adversarial examples around each sample, replacing the the perturbation set $B(x,\epsilon)$.
From a different view, \citeauthor{pmlr-v119-stutz20a}~(\citeyear{pmlr-v119-stutz20a}) suggested calibrating the confidence scores during adversarial training.

Though significant, the generalization problem of adversarial training on unseen attacks is only occasionally studied at this time.
One possible reason is that our understanding of adversarial examples is limited and incomplete.
The truth of adversarial examples is still underground, which also needs much effort.

\section{Conclusion and Future Directions}
In this paper, we reviewed the current adversarial training methods for adversarial robustness.
To our best knowledge, for the first time, we review adversarial training with a novel taxonomy and discuss the generalization problem in adversarial training.
We also summarize the benchmarks and provide performance comparisons of different methods.
Despite extensive efforts, the vulnerability of deep learning models to adversarial examples haven't been solved by adversarial training.
Several open problems remain yet to solve, which are summarized as follows.

\noindent \textbf{Min-Max Optimization in Adversarial Training.}
Adversarial training is formulated as a min-max problem.
However, due to the non-convexity of deep neural networks, it is very challenging to obtain the global optimum for adversarial training.
In existing methods, PGD is a prevalent technique for approximating the optimum, as \citeauthor{DBLP:conf/iclr/MadryMSTV18}~(\citeyear{DBLP:conf/iclr/MadryMSTV18}) empirically proved the tractability of adversarial training with PGD.
But it can hardly provide a ``robustness certificate" after solving the problem~\cite{razaviyayn2020nonconvex}.
In other words, the robustness of adversarially trained models is not guaranteed~\cite{kang2019transfer}.
For this purpose, the development of new techniques for solving non-convex min-max problems is necessary and crucial. 

\noindent \textbf{Overfitting in Adversarial Training.}
Overfitting is an old topic in deep learning, and there are effective countermeasures to alleviate the overfitting.
But, in adversarial training, overfitting seems to be more severe and those common techniques used in deep learning help little~\cite{rice2020overfitting}.
The generalization gap between adversarial training accuracy and testing accuracy is very large.
From the perspective of generalization, the theory of sample complexity~\cite{NIPS2018_7749} explains such a phenomenon partially, and is supported by experimental results in derivative works~\cite{NIPS2019_9388,NIPS2019_9298}.
As suggested by~\cite{NIPS2018_7749}, it is essential to explore the intersections between robustness, classifiers and data distribution.
Some open problems can be found in~\cite{NIPS2018_7749}.

\noindent \textbf{Beyond Adversarial Training.}
Though many theories have been proposed for improving adversarial training, it is undeniable that these improvements are less effective than claimed~\cite{pang2020bag}.
Some basic settings \eg training schedule, early stopping, seem to owe much on the improvement of adversarial training.
The shreds of evidence in~\cite{yang2020closer,Stutz_2019_CVPR} show that adversarial training might not be the optimal solution for obtaining models with desirable robustness and accuracy.
The trade-off between robustness and generalization can also be seen as an intrinsic limitation of adversarial training.
Thus it is critical and necessary to investigate new methods beyond adversarial training for adversarial robustness in the future.

\bibliographystyle{named}
\bibliography{ijcai20}

\end{document}